\documentclass{article}
\usepackage{arxiv}

\usepackage{cite}
\usepackage{amsmath,amssymb,amsfonts}
\usepackage{algorithmic, algorithm}
\usepackage{graphicx}
\usepackage{textcomp}
\usepackage{xcolor}

\usepackage{todonotes}
\usepackage{subcaption}
\usepackage{textcomp}
\usepackage{transparent}
\usepackage{makecell}
\usepackage{multirow}
\usepackage{color, colortbl}
\usepackage{xpatch}
\usepackage{longtable}
\usepackage{lipsum}
\usepackage{array}
\usepackage[colorlinks=true, allcolors=black, pdftitle={XPROB}]{hyperref}

\newcolumntype{x}[1]{>{\centering\arraybackslash}p{#1}}

\def\BibTeX{{\rm B\kern-.05em{\sc i\kern-.025em b}\kern-.08em
    T\kern-.1667em\lower.7ex\hbox{E}\kern-.125emX}}

\DeclareMathOperator*{\argmax}{arg\,max}



\makeatletter
\xpatchcmd{\@thm}{\thm@headpunct{.}}{\thm@headpunct{}}{}{}
\makeatother

\newcommand\pbased{XPROB}
\newcommand\gbased{XPROAX}

\newcommand{\hlBlue}[2][0.1]{{\transparent{#1}\colorbox{blue}{\transparent{1}#2}}}
\newcommand{\hlRed}[2][0.1]{{\transparent{#1}\colorbox{red}{\transparent{1}#2}}}

\newcommand\Tstrut{\rule{0pt}{2.1ex}}       

\renewcommand{\algorithmicrequire}{\textbf{Input:}}
\renewcommand{\algorithmicensure}{\textbf{Output:}}
\definecolor{Gray}{gray}{0.9}

\newcolumntype{C}[1]{>{\centering}m{#1}}
\renewcommand{\shorttitle}{Transparent Neighborhood Approximation for Text Classifier Explanation}

\begin{document}

\title{
Transparent Neighborhood Approximation for Text Classifier Explanation by Probability-based Editing
}

\author{
\makebox[.3\linewidth]{Yi~Cai} \\
Dept. of Math. and Comp. Science \\
\textit{Freie Universität Berlin}\\
Berlin, Germany \\
yi.cai@fu-berlin.de
\And
\makebox[.3\linewidth]{Arthur~Zimek}\\
Dept. of Math. and Comp. Science \\
\textit{University of Southern Denmark}\\
Odense, Denmark \\
zimek@imada.sdu.dk
\And
\makebox[.3\linewidth]{Eirini~Ntoutsi}\\
Research Institute CODE \\
\textit{Universität der Bundeswehr München}\\
Munich, Germany \\
eirini.ntoutsi@unibw.de
\And
\makebox[.3\linewidth]{Gerhard~Wunder}\\
Dept. of Math. and Comp. Science \\
\textit{Freie Universität Berlin}\\
Berlin, Germany \\
gerhard.wunder@fu-berlin.de
}
\maketitle

\begin{abstract}
    Recent literature highlights the critical role of neighborhood construction in deriving model-agnostic explanations, with a growing trend toward deploying generative models to improve synthetic instance quality, especially for explaining text classifiers.
    These approaches overcome the challenges in neighborhood construction posed by the unstructured nature of texts, thereby improving the quality of explanations.
    However, the deployed generators are usually implemented via neural networks and lack inherent explainability, sparking arguments over the transparency of the explanation process itself.
    To address this limitation while preserving neighborhood quality, this paper introduces a probability-based editing method as an alternative to black-box text generators.
    This approach generates neighboring texts by implementing manipulations based on in-text contexts.
    Substituting the generator-based construction process with recursive probability-based editing, the resultant explanation method, \pbased{}~(explainer with probability-based editing), exhibits competitive performance according to the evaluation conducted on two real-world datasets.
    Additionally, \pbased{}'s fully transparent and more controllable construction process leads to superior stability compared to the generator-based explainers.
\end{abstract}



\section{Introduction}
Advances in machine learning, particularly deep learning, have enabled applications in various real-world scenarios~\cite{pouyanfar2018survey}.
Accompanying improvements in performance, the increasing model complexity and the exploding number of parameters~\cite{bernstein2021freely} prohibit human users from comprehending the decisions made by these data-driven models, as the decision rules are implicitly learned from the data presented.
The absence of reasoning for model decisions keeps raising concerns about the transparency of AI-driven systems~\cite{goodman2017european}.
Encouraged by these concerns, extensive research in explainable AI (XAI) emerges, intending to uncover the decision-making process hiding under the machine learning black boxes.

A major focus in enhancing explainability is on identifying evidence supporting model decisions.
Known as attribution methods, these approaches assign scores to individual input features of a specific decision, referred to as the \textit{explicand}.
These attribution scores quantify each feature's contribution to the target decision.
A common solution for deriving feature attribution involves analyzing the correlation between input features and model outcomes~\cite{carvalho2019machine}.
Requiring only query-level access to the target model, this kind of method delivers explanations in a post-hoc manner without delving into model internal details.
Query-level access means that the to-be-explained model can only be interacted with through its input and output interfaces.
Making minimal, if any, assumptions about the model being explained, these solutions are termed model-agnostic approaches.

In the explanation process, a model-agnostic approach begins by creating a synthetic set (termed the \textit{neighborhood}) containing instances surrounding the explicand~\cite{ribeiro2016should, lundberg2017unified}.
These instances are then labeled with outputs from the target black box, providing pseudo labels that reflect the model's behaviors.
During the synthetic set construction, the locality constraint~\cite{laugel2018defining} emphasizes the spatial closeness of the generated instances to the explicand, ensuring that the derived explanation concentrates exclusively on model behaviors relevant to the inquired decision.
Once the neighborhood is determined, explanatory information is indirectly extracted by investigating the impacts of altered features on model outcomes.

The quality of generated instances, which are the sole reflections of model behaviors, significantly affects the performance of model-agnostic explainers.
Although neighborhood construction appears relatively simple for structured data, such as tabular~\cite{laugel2018defining} and image~\cite{petsiuk2018rise} data, solving the same task for textual data is nontrivial.
The strong inherent feature interaction in texts degrades the effectiveness of straightforward generation approaches like random word dropping~\cite{ribeiro2016should}.
A recent line of works~\cite{cai2021xproax, lampridis2023explaining} proposes using generative models for neighborhood construction, which produces realistic neighboring texts -- semantically meaningful and grammatically correct -- thereby enhancing explanation quality.
However, the deployment of non-self-explainable components introduces additional opacity to the constructed neighborhoods and, consequently, to the explanations derived from them.

In response to transparency concerns in the explanation procedure, this paper proposes \pbased{}\footnote{Source code is available at: \url{https://https://github.com/caiy0220/XPROB}.}~(e\underline{x}plainer with \underline{prob}ability-based editing), an approach generating realistic texts for neighborhood construction through editing guided by local \textit{n}-gram contexts.
Our experiments on two real-world datasets empirically show the competitive performance of the proposed explainer without relying on black box generators.
The experimental results also validate concerns about the inherent opacity of generator-based explainers by highlighting their complicated dependencies on generative models.
Furthermore, \pbased{}'s neighborhood construction is more controllable as its probability-based editing is a deterministic process following explicit rules, contrasting with generator-based solutions, where the effects of latent space perturbations on generated texts are unforeseeable.


\section{Related Work}\label{sec:relatedWork}
Constructing neighborhoods is a fundamental task in deriving model-agnostic explanations.
The earlier attempts to accomplish the task can be summarised as input perturbation~\cite{vstrumbelj2009explaining}, which is well-suited for structured data.
This approach generates neighboring instances by altering the feature values of an explicand.
Specifically, input perturbation applies slight fluctuation in the vicinity of a given input for numeric features~\cite{laugel2018defining} or samples categorical feature values from some empirical distribution~\cite{chen2023algorithms}.
While the strategy appears straightforward, perturbing input adheres to the locality constraint.

Building on the concept of input perturbation, LIME~\cite{ribeiro2016should} constructs neighborhoods of text by randomly dropping words from an explicand, repeating the process until sufficient instances are collected.
Similarly, input occlusion~\cite{jin2019towards} perturbs text by replacing words with padding tokens.
Although both methods generate synthetics similar to the original explicand, they face limitations with textual data due to strong inherent feature interactions.
While adhering to the locality constraint, the randomly perturbed texts can drive the concentration of explanations away from the natural data manifold~\cite{chen2020true, ghalebikesabi2021locality}, potentially leading to explanations that do not accurately represent model behaviors~\cite{frye2020shapley}.
Besides, the feasibility of these methods is constrained by the length of the target text.
For explicands with only a few words, input perturbation struggles to gather sufficient unique samples for further analysis.


XSPELLS~\cite{lampridis2020explaining, lampridis2023explaining} addresses the limitations of input perturbation by deploying a variational autoencoder~\cite{bowman2016generating} for textual neighborhood construction.
This generator-based explainer samples instances in a numeric latent space maintained by the generative autoencoder, selecting neighbors based on their proximity to the latent representation of the explicand.
The sampled latent vectors are then decoded into texts by the generator for succeeding explanation processes. 
A follow-up work, XPROAX~\cite{cai2021xproax}, refines the latent space sampling with a two-staged progressive approximation.
Specifically, it employs iterative latent space interpolations between points with distinct labels, which effectively highlights decision boundaries relevant to the queried decision.
XPROAX also specifies the locality-preserving and reconstruction properties as key criteria in selecting generative models for model-agnostic explanations.
Prior applications of the generator-based construction strategy include ABELE~\cite{guidotti2019black}, which explains image classifiers by refining embedding sampling with a genetic algorithm, and methods for tabular data classifiers~\cite{shankaranarayana2019alime, frye2020shapley}, which aims at aligning the distribution of generated instances with the data manifold.
Moreover, recent research extends the use of generators to craft counterfactuals~\cite{feder2021causalm, rodriguez2021beyond}, providing an alternative form of explanation that explores minimal changes needed to alter the original outcome.

Despite the popularity of generators in tasks related to explainability, the heavy reliance of generator-based explainers on these opaque components undermines the transparency of the explanations derived, which requires further discussion.
In fact, explaining generative models remains an open question~\cite{longo2024explainable}, particularly challenging due to the complex nature of their outputs, such as texts and images. 
Moreover, generative models generally require more information than classifiers to operate on the same data manifold.
This is evidenced by the fact that generative models can be more readily fine-tuned for classification tasks~\cite{brown2020language}, but not the other way around, implying that generators often possess capabilities beyond classifiers.
Consequently, generator-based attribution methods risk complicating the explanation process by explaining one black box with another, potentially more complicated one.

\section{Explaining with Probability-based Editing}\label{sec:xproaxStats}
Given a text classifier $f(\cdot)$ and an explicand consisting of a sequence of tokens $\boldsymbol{x}=(w_1, w_2, \ldots, w_n)$, the objective of feature attribution methods is to determine a vector $\boldsymbol{\xi}$, whose elements indicate the contributions of input features to the decision\footnote{For simplicity, we assume a binary classification problem: $y\in \{+, -\}$.} $y=\argmax f(\boldsymbol{x})$.
To extract explanatory information under query-level access, \pbased{}~constructs a neighborhood $N(\boldsymbol{x})$ around the explicand, then quantifies feature contributions indirectly using a surrogate $g(\cdot)$ (details about the extraction are presented in Section~\ref{sec:extraction}).
The surrogate model $g(\cdot)$ is a self-explainable weak learner, trained on the pseudo-labeled neighborhood, and designed to emulate the local behavior of the black box model $f(\boldsymbol{x})$.
Consistent with prior research~\cite{alvarez2017causal}, neighboring texts are defined as texts sharing similar components and meanings.

\subsection{Probability-based Editing}\label{sec:probEdit}
While input perturbation fails to generate synthetics representing the underlying data manifold of a classification task, adopting generative models produces realistic neighboring texts.
This can be formally described as creating a synthetic set where each element $\forall~\boldsymbol{\dot{x}}\in N(\boldsymbol{x})$ satisfies $\boldsymbol{\dot{x}}\sim \mathcal{X}$, with $\mathcal{X}$ denoting the target text domain.
A text generator is trained to learn the conditional probability distribution, allowing it to extend a given prompt with the most likely word:
\begin{equation*}
    w_{l+1}=\argmax_{w}P(w|w_1,w_2, \ldots, w_l)
\end{equation*}
Given that text generators function as probability distribution estimators, we propose an editing approach driven directly by empirical probability distributions as a transparent alternative.

Different from the objective of extending given prompts, text manipulation for neighborhood construction involves more operation than appending tokens to existing texts and should align with the locality constraint.
Therefore, we redefine the editing task to identify the optimal manipulation operation that integrates a selected token from the explicand $w\in\boldsymbol{x}$ into a prototype $\boldsymbol{\hat{x}}=(\hat{w}_1,\hat{w}_2, \ldots, \hat{w}_l)$.
The integration maximizes the probability of the resultant text:
\begin{equation} \label{eq:edit}
    \mathcal{O}(w,\boldsymbol{\hat{x}}) = \argmax_{1\leq i < j \leq l} P(\hat{w}_1, \ldots, \hat{w}_i, w, \hat{w}_j, \ldots, \hat{w}_l)
\end{equation}
where the probability $P(\hat{w}_1, \ldots, \hat{w}_i, w, \hat{w}_j, \ldots, \hat{w}_l)$ is empirically estimated on a corpus $X_p\sim \mathcal{X}$ sampled from the target text domain.
The manipulation operation, defined by the indices $i$ and $j$, refers to either inserting at position $i$ of the prototype when $j=i+1$, or replacing a contiguous subsequence $\boldsymbol{\hat{x}}_{i+1}^{j-1}=(\hat{w}_{i+1}, \ldots, \hat{w}_{j-1})$ in $\boldsymbol{\hat{x}}$ (denoted by $\boldsymbol{\hat{x}}_{i+1}^{j-1}\prec\boldsymbol{\hat{x}}$) otherwise.
A prototype, which is a neighboring counterfactual, is selected from the retained corpus $X_p$ (see further details about prototype selection in Section~\ref{sec:neighborhood}).
By recursively integrating explicand components to prototypes, \pbased{}~produces a series of neighboring texts that share components with $\boldsymbol{x}$, performing a gradual transition from one class (the label assigned to $\boldsymbol{\hat{x}}$) to another (the label of $\boldsymbol{x}$) that highlights the target decision boundary.

Implementing manipulation as outlined in Eq.~\ref{eq:edit} ensures the grammatical correctness of the generated texts. However, using the entire sequence as context restricts the process to only reproducing texts present in $X_p$.
To enable exploration beyond existing token combinations, the editing process considers local \textit{n}-gram contexts~\cite{jurafsky2000speech} instead of whole sequences.
Specifically, the \textit{n}-gram context consists of the preceding sequence $\boldsymbol{\hat{x}}^i_{i-n+1}$ and the succeeding sequence $\boldsymbol{\hat{x}}^{j+n-1}_j$ for a manipulation operation specified by $i$ and $j$. 
Updating the editing objective with the local \textit{n}-gram context yields:
\begin{equation*}
    \mathcal{O}(w,\boldsymbol{\hat{x}}) = \argmax_{1\leq i < j \leq l}P_{\mathit{pre}}(w|\boldsymbol{\hat{x}}^{i}_{i-n+1})P_{\mathit{suc}}(w|\boldsymbol{\hat{x}}^{j+n-1}_j)
\end{equation*}
The optimal operation can be identified in quadratic time w.r.t.\ the prototype length.
Denoting word sequence concatenation as $\boldsymbol{x}^b_a + \boldsymbol{x}^d_c = (w_a, \ldots, w_b, w_c, \ldots, w_d)$, the conditional probability of integrating $w$ given the \textit{preceding} and \textit{succeeding} words can be written as:
\begin{align*}
    P_{\mathit{pre}}(w|\boldsymbol{\hat{x}}^i_{i-n+1})&=\frac{P(\boldsymbol{\hat{x}}^i_{i-n+1} + (w))}{P(\boldsymbol{\hat{x}}^i_{i-n+1})} \\
    P_{\mathit{suc}}(w|\boldsymbol{\hat{x}}^{j+n-1}_j)&=\frac{P((w) + \boldsymbol{\hat{x}}^{j+n-1}_j)}{P(\boldsymbol{\hat{x}}^{j+n-1}_j)}
\end{align*}
Albeit not explicitly given, the above conditional probabilities can be estimated using the retained corpus $X_p$:
\begin{align*}
    P_{\mathit{pre}}(w|\boldsymbol{\hat{x}}^{i}_{i-n+1})=&~\frac{|\{\boldsymbol{x}|\boldsymbol{x}\in X_p, (\boldsymbol{\hat{x}}^{i}_{i-n+1} + (w))\prec \boldsymbol{x}\}|}{|\{\boldsymbol{x}|\boldsymbol{x}\in X_p, \boldsymbol{\hat{x}}^{i}_{i-n+1}\prec \boldsymbol{x}\}|} \\
    P_{\mathit{suc}}(w|\boldsymbol{\hat{x}}^{j+n-1}_{j})=&~\frac{|\{\boldsymbol{x}|\boldsymbol{x}\in X_p, ((w) + \boldsymbol{\hat{x}}^{j+n-1}_{j}) \prec \boldsymbol{x}\}|}{|\{\boldsymbol{x}|\boldsymbol{x}\in X_p, \boldsymbol{\hat{x}}^{j+n-1}_{j}\prec \boldsymbol{x}\}|}
\end{align*}
For subsequences $\boldsymbol{\hat{x}}^{i}_{i-n+1} + (w)$ absent in $X_p$, we assign a minimal value $\epsilon={(|X_p|+1)}^{-1}$ instead of $0$ to avoid the probability collapsing to zero during the multiplication of $P_{\mathit{pre}}$ with $P_{\mathit{suc}}$.
Given the minimal value, we set the threshold for a valid edit at $\epsilon^2$.
This threshold value implies that an edit is unfeasible if the target token $w$ never appears in the context formed by either the preceding or succeeding texts according to $X_p$.
The choice of $n$, which determines the range of local contexts, balances between semantic correctness and feasibility of the edit.

To allow manipulation at the beginning or the end of a text, \textit{n} padding tokens \textlangle\textit{PAD}\textrangle~are concatenated to both the head and the tail of a prototype. 
However, this introduces a bias towards the padding tokens due to their omnipresence, which inadvertently induces the manipulation process to replace an entire text sequence with $w$, considering only the padding tokens as the local context.
Such an abrupt change, which replaces a whole sentence, contradicts the intention of implementing gradual modifications.
To address this issue, a penalty based on edit length is incorporated into the manipulation objective:
\begin{equation} \label{eq:contrainedEdit}
    \mathcal{O}(w,\boldsymbol{\hat{x}}) = \argmax_{1\leq i < j \leq l}\frac{P_{\mathit{pre}}(w|\boldsymbol{\hat{x}}^{i}_{i-n+1})P_{\mathit{suc}}(w|\boldsymbol{\hat{x}}^{j+n-1}_j)}{e^{j-i}}
\end{equation}
This soft constraint on edit length encourages operations that affect fewer prototype components.

\subsection{Neighborhood Approximation with Recursive Editing}\label{sec:neighborhood}
Guided by probability-based editing, \pbased{}~builds neighborhoods through recursive text manipulations.
The process is initialized with real counterfactuals from $X_p$ that serve as prototypes $\boldsymbol{\hat{x}}$, since the manipulation formulated by Eq.~\ref{eq:contrainedEdit} requires local \textit{n}-gram contexts and does not generate texts from scratch.
By integrating explicand components into counterfactuals, \pbased{}~implements a gradual transition in model predictions from one class to another, resulting in synthetics that reveal the decision boundary situated in between.

To ensure the locality of the generated texts, prototype selection follows their spatial closeness to the explicand.
Specifically, the proximity of text pairs is measured by the cosine distance between their \textit{tf-idf} vectors~\cite{manning2008introduction} with a vectorizer fitted on $X_p$.
The distance measure reflects text similarity in terms of components, aligning with the definition of neighboring texts~\cite{alvarez2017causal}.
The texts closest to $\boldsymbol{x}$ form the initial prototype set $S$.
Although there are other text similarity measures, such as the distance of text embeddings~\cite{reimers2019sentence}, \textit{tf-idf} is preferred in this context to avoid introducing any external opacity to the explanation framework.

Taking as input an explicand $\boldsymbol{x}$ and the corresponding prototype set $S$, the detailed neighborhood construction process is outlined in Algorithm~\ref{alg:pbEdit}.
For each word-prototype pair $(w, \boldsymbol{\hat{x}})$ (line 3), the recursive process attempts probability-based editing to produce a synthetic text $\boldsymbol{\dot{x}}$ (lines 4, 5).
Manipulation may be skipped if the prototype already contains $w$ or if the optimal probability obtained by editing does not exceed the validity threshold $\epsilon^2$.
The synthetics produced in one iteration are used to update $S$ for the subsequent round, enabling recursive editing (lines 6, 8).
Throughout the iterations, all intermediate texts are collected to shape the final neighborhood $N(\boldsymbol{x})$ (line 9).
The generation process terminates when there are no feasible operations on the updated prototype set or when the neighborhood has collected sufficient instances.

\begin{algorithm}[tb]
\caption{Recursive probability-based Edit}\label{alg:pbEdit}
\begin{algorithmic}[1]
\renewcommand{\algorithmicrequire}{\textbf{Input:}}
\renewcommand{\algorithmicensure}{\textbf{Output:}}
\REQUIRE $\boldsymbol{x}$: explicand; $S$: prototype set
\ENSURE  $N(\boldsymbol{x})$: neighborhood of $\boldsymbol{x}$;
\WHILE{terminate criterion NOT met}
    \STATE $N(\boldsymbol{x})=S$, $S_{\mathit{new}}=\varnothing$
    \FOR{$w \in \boldsymbol{x}$, $\boldsymbol{\hat{x}}\in S$}
        \STATE $i,j=\textrm{Edit}(w, \boldsymbol{\hat{x}})$
        \STATE $\boldsymbol{\dot{x}}=(\hat{w}_1, \ldots, \hat{w}_i, w, \hat{w}_j, \ldots, \hat{w}_l)$
        \STATE $S_{\mathit{new}}.\textrm{add}(\boldsymbol{\dot{x}})$
    \ENDFOR
    \STATE $S=S_{\mathit{new}} - S_{\mathit{new}} \cap N(\boldsymbol{x})$  
    \STATE $N(\boldsymbol{x})=N(\boldsymbol{x})\cup S_{\mathit{new}}$    
\ENDWHILE
\RETURN $N(\boldsymbol{x})$ 
\end{algorithmic} 
\end{algorithm}

\subsection{Extraction of Local Explanations}\label{sec:extraction}
Once neighborhood construction is completed, explanatory information is extracted using a linear regression model $g(\cdot)$, trained on $N(\boldsymbol{x})$ with a weighted square loss:
\begin{equation*}
\mathcal{L}(N(\boldsymbol{x}), \boldsymbol{x})=\sum_{\boldsymbol{\dot{x}}\in N(\boldsymbol{x})}\exp\left(\frac{-{d(\boldsymbol{\dot{x}}, \boldsymbol{x})}^2}{\sigma^2}\right)\cdot{(f(\boldsymbol{\dot{x}})-g(\boldsymbol{\dot{x}}))}^2
\end{equation*}
where each synthetic is weighted by its cosine distance to the explicand $d(\boldsymbol{\dot{x}}, \boldsymbol{x})$, further strengthening locality.
The surrogate mimics the behaviors of $f(\cdot)$ as reflected by $N(\boldsymbol{x})$, serving as a proxy for explanation extraction.
Specifically, the coefficients determined by $g(\cdot)$ quantify the attributions to input features.
Since the novel neighborhood construction introduces external words not originally part of the explicand, the derived feature attributions can be split into two parts: \textit{intrinsic} and \textit{extrinsic} attributions.
While intrinsic attributions reveal feature contributions to the target decision, influential extrinsics indicate important observations for inference in the target context, thereby facilitating the explanation for a decision~\cite{ma2017salient, wang2011image}.

In addition to feature attributions, extracting factual and counterfactual instances from the more realistic neighboring texts forms instance-level explanations $\boldsymbol{\xi}^*$, which complement word-level explanations $\boldsymbol{\xi}$.
Analyzing the intersections and distinctions between factuals and counterfactuals helps to highlight the relevant components of a prediction.
Contrary to greedily selecting instances closest to $\boldsymbol{x}$, which often produces redundant samples, the construction of $\boldsymbol{\xi}^*$ incorporates diversity~\cite{gong2019diversity} alongside spatial closeness.
Over an iterative selection process, $\boldsymbol{\xi}^*$ sequentially absorbs the best-fitting instance described by:
\begin{equation*}
    \boldsymbol{\dot{x}} = \argmax_{\boldsymbol{\dot{x}}\in N(\boldsymbol{x})} \lambda(1-d(\boldsymbol{\dot{x}}, \boldsymbol{x}))+(1-\lambda) \sum_{\boldsymbol{\dot{x}'}\in \boldsymbol{\xi}^*}\frac{d(\Delta \boldsymbol{\dot{x}}, \Delta \boldsymbol{\dot{x}'})}{|\boldsymbol{\xi}^*|} 
\end{equation*}
where $\Delta \boldsymbol{\dot{x}}=\boldsymbol{\dot{x}}-\boldsymbol{x}$ denotes the manipulation direction from the explicand to a generated text.
While the first term measures the spatial closeness, the second term quantifies the diversity of a candidate relative to the previously selected instances.
Balanced by the parameter $\lambda\in[0,1]$, the selection process avoids redundancy in $\boldsymbol{\xi}^*$.
\section{Evaluation}\label{sec:evaluation}
To evaluate the quality of explanations derived by adopting probability-based editing, this section tests \pbased{}~on two real-world datasets with both qualitative (Section~\ref{sec:qualitative}) and quantitative (Section~\ref{sec:quantitative}) assessments.
Subsequently, an experiment designed to verify explanation stability is conducted (Section~\ref{sec:stability}).
The final part of the experiments explores the impact of external resources on the quality of explanations, which validates concerns about leveraging opaque generators in the context of neighborhood construction (Section~\ref{sec:dependency}).

\subsection{Experimental setup}\label{sec:exp_setup}
\emph{Datasets}: 
Two real-world datasets, Yelp reviews~\cite{shen2017style} and Amazon reviews polarity~\cite{mcauley2013hidden}, are utilized for the evaluation of explanations.
The \textit{Yelp} dataset comprises restaurant reviews, labeled as either positive or negative.
The \textit{Amazon review polarity} dataset includes customer reviews of products with binarized labels.
For both datasets, three disjoint subsets are created for classifier training: training, validation, and test sets with sizes of \textit{20k}/\textit{2k}/\textit{4k}, respectively.
Additionally, each dataset contains \textit{200k} unlabeled entries, which serve as a corpus reflecting text distribution for the training of generative models used by generator-based explainers.
\pbased{}, which also utilizes an external corpus, downsamples \textit{20k} texts from the unlabeled set to form $X_p$ for probability distribution estimation and prototype selection.

\emph{Text classifiers}: 
In principle, model-agnostic explainers are applicable to classifiers regardless of their underlying architecture.
For the sake of evaluation, we test the explainers on two black box models: a BERT model~\cite{kenton2019bert} and an eight-layer LSTM network~\cite{hochreiter1997long}, which represent popular architectures for natural language processing.
These models are well-suited as black boxes due to their complexity and non-linearity.
We fine-tuned a publicly available pretrained version of BERT\footnote{https://huggingface.co/docs/transformers/model\_doc/bert} on each dataset.
The LSTM features an architecture starting with three dense layers that learn word representations, followed by an LSTM layer capturing the sequential information of texts.
The output of the LSTM layer is then processed through another four additional dense layers, including an output layer for the classification tasks concatenated in the end.
Table~\ref{tbl:dataset} presents the accuracy of the trained classifiers on both datasets.

\begin{table}[tbp]
\caption{Details about datasets and black box models}
\centering
\begin{tabular}{|c|C{1.1cm}|C{1.1cm}|C{1.1cm}|c|c|} 
\hline
\multirow{2}{*}{Dataset} & \multicolumn{3}{c|}{Black box} & \multicolumn{2}{c|}{Accuracy$^{\mathrm{a}}$} \\
\cline{2-6}
 & & & & & \\ [-2.7mm]
 & Training & Valid & Test & BERT & LSTM \\ 
\hline
Amazon & 20K & 2K & 4K & 0.8543 & 0.7795 \\ 
\hline
Yelp & 20K & 2K & 4K & 0.9783 & 0.9548 \\
\hline
\multicolumn{6}{l}{$^{\mathrm{a}}$Performances are reported on the test sets.}
\end{tabular}
\label{tbl:dataset}
\end{table}

\emph{Competitors}: 
As this paper focuses on improving the neighborhood construction process and providing an alternative to generator-based solutions, we compare \pbased{}~with one approach that adopts input perturbation and three generator-based explainers:
\begin{itemize}
    \item \textbf{LIME}~\cite{ribeiro2016should}: a local explanation method constructing neighborhoods by word dropping.
    \item \textbf{XSPELLS}~\cite{lampridis2020explaining}: a text-classifier-specified explainer generating neighbors through random latent space sampling and deriving explanations from a latent decision tree.
    \item \textbf{ABELE}~\cite{guidotti2019black}: a generator-based explanation method constructing neighborhoods with a heuristic approach.
    \item \textbf{XPROAX}~\cite{cai2021xproax}: an explainer implementing progressive neighborhood approximation through a generative model.
\end{itemize}
We note that ABELE~\cite{guidotti2019black} is designed to explain image classifiers, but it can be readily adapted to deal with textual data since neighborhood construction in the latent space is independent of the original input formats.

\emph{Generative model}:
Following the suggestion by~\cite{cai2021xproax}, the generator-dependent competitors, namely XSPELLS, ABELE, and XPROAX, adopt a DAAE~\cite{shen2020educating} as their generator to ensure the locality of the generated texts.
The DAAE, a symmetric generative autoencoder, is designed to learn a locality-preserving latent space.
The hyperparameters for the generator are configured according to the original paper~\cite{shen2020educating}.
The autoencoder has a bottleneck of size 128 with both its encoder and decoder implemented by a one-layer bidirectional LSTM with 1024 units. 

\emph{Hyperparameters for \pbased{}}:
Across all test cases, the neighborhood population $p$ and the number of initial prototypes $k$ are set to 400 and 80 for \pbased{}\footnote{The studies on hyperparameter impacts are not presented here due to the space limitation.}.
Given the relatively shorter review texts, unigram ($n=1$) is employed for representing local contexts, allowing more feasible edits during neighborhood construction.

\subsection{Qualitative evaluation}\label{sec:qualitative}
For qualitative evaluation, Table~\ref{tbl:quality} lists sample explanations from different approaches for the selected decisions.
In this table, feature attribution is visualized via saliency maps, where a blue (red) background indicates a contribution to positive (negative, respectively) sentiment.
The intensity of the background color corresponds to the magnitude of feature attribution.
The visualization is not applied to XSPELLS for a distinguishing purpose.
Rather than estimating feature attributions, XSPELLS counts relative word frequencies (numbers in brackets) on the branch of the latent decision tree associated with the explicand.
Beyond feature attributions, the explainers, except LIME, deliver additional explanatory information that further reveals model behaviors. 
This includes the most influential \textit{extrinsic words} as well as the selected factual and counterfactual examples.
For the extrinsic words, only the ones with distinctly high attribution scores are reported.

\begin{table*}[tbp]
\centering
\fontsize{7.5}{8.5}\selectfont
\caption{Example explanations by different methods}
\begin{tabular}{c | p{6.35cm} p{7.65cm} } 
    \hline
    \rowcolor{Gray}
    \textbf{Input 1} & \multicolumn{2}{p{14.4cm}}{excellent sta for beginning french students. \hfill Dataset: \textbf{Amazon}, \textbf{BERT} $f(\cdot)$: {\textcolor{blue}{0.99}}\Tstrut}
    \\ \hline
    \\[-0.5em] \textbf{LIME} & 
    \multicolumn{2}{l}{\textbf{Saliency}: \hlBlue[0.22]{excellent} sta for beginning french students.}
    \\[0.5em] \hline \\[-1em] 
    \multirow{2}{*}[-3.5em]{\textbf{XSPELLS}} &
    \textbf{Factuals}:\Tstrut & \textbf{Counterfactuals}:\Tstrut \\
    & 1) excellent for the $\langle unk\rangle^{\mathrm{b}}$ $\langle unk\rangle$! & 1) chicken for ground deep $\langle unk\rangle$\\
    & 2) an essential watch for the world coach students! & 2) excellent piece of the sta and mediocre yoga program as well \\
    & 3) excellent summer for the field bible! & 3) great beats see tv show cannot.\\
    & 4) if not for this price , then the french is for you & 4) an empty direction for christian\\
    & 5) excellent sta for the preface. & 5) worst shark move next book ever! \\
    & \textbf{Most frequent words in factuals}: & \textbf{Most frequent words in counterfactuals}:\\
    & excellent (0.114), best (0.086), students (0.057) & ground (0.024), sta (0.024), chicken (0.024)
    \\[0.2em] \hline \\[-1em] 
    \multirow{2}{*}[-3em]{\textbf{ABELE}} & 
    \textbf{Saliency}: \hlBlue[0.34]{excellent} sta for beginning french students. &
    \textbf{Extrinsic words}: \hlRed[0.36]{please}~\hlRed[0.16]{not}\\
    & \textbf{Factuals}: & \textbf{Counterfactuals}: \\
    & 1) excellent sta for beginning french philosophy. & 1) please sta for the french courses.\\
    & 2) excellent sta for beginning french philosophy. & 2) please sta for the french courses.\\
    & 3) excellent sta for beginning french philosophy. & 3) please sta for the french courses.\\
    & 4) excellent sta for beginning french philosophy. & 4) please sta for the french courses.\\
    & 5) excellent sta for beginning french philosophy. & 5) please sta for the french courses.
    \\[0.2em] \hline
    \\[-1em] \multirow{2}{*}[-3em]{\textbf{\gbased{}}} & 
    \textbf{Saliency}: \hlBlue[0.48]{excellent} sta for beginning french students. &
    \textbf{Extrinsic words}: \hlRed[0.38]{disappointing}~\hlRed[0.35]{worst}\\
    & \textbf{Factuals}: & \textbf{Counterfactuals}: \\
    & 1) excellent sta for beginning french philosophy. & 1) a disappointing for beginning to $\langle unk\rangle$ philosophy\\
    & 2) excellent sta for beginning to travel students. & 2) excellent sta bad beginning and cover $\langle unk\rangle$ waste of\\
    & 3) excellent action for beginning of ed. & 3) the sta to the bible on philosophy?\\
    & 4) excellent sta for beginning to craft physics. & 4) not woh the price nothing to deceiving.\\
    & 5) excellent sta for beginning to craft students. & 5) bad end to the series...
    \\[0.2em] \hline
    \\[-1em] \multirow{2}{*}[-3em]{\textbf{\pbased{}}} & 
    \textbf{Saliency}: \hlBlue[0.33]{excellent} sta for beginning french students. &
    \textbf{Extrinsic words}: \hlRed[0.37]{ridiculous} \hlRed[0.18]{but}\\
    & \textbf{Factuals}: & \textbf{Counterfactuals}: \\
    & 1) sta for students & 1) a ridiculous sta to french students \\
    & 2) sta for beginning french students. & 2) excellent sta for a reference but bad for the beginning french students\\
    & 3) excellent a beginning french students. & 3) beginning a ridiculous sta to a french students. \\
    & 4) excellent sta for a french students & 4) very average and for the beginning of the french students \\
    & 5) excellent sta for beginning french & 5) excellent sta for good beginning... weak finish
    \\[0.2em] \hline
    \multicolumn{3}{l}{$^{\mathrm{b}}$The generic token for words out of vocabulary.} \\
    \multicolumn{3}{l}{}\\[-0.3em] \hline
    
    \rowcolor{Gray}
    \textbf{Input 2} & \multicolumn{2}{p{14.4cm}}{the desserts were very bland. \hfill Dataset: \textbf{Yelp}, \textbf{LSTM} $b(\cdot)$: {\textcolor{red}{0.99}}\Tstrut} 
    \\ \hline
    \\[-0.5em] \textbf{LIME} & 
    \multicolumn{2}{l}{\textbf{Saliency}: the desserts were very \hlRed[0.03]{bland}.}
    \\[0.5em] \hline \\[-1em] 
    \multirow{2}{*}[-3.5em]{\textbf{XSPELLS}}
    & \textbf{Factuals}:\Tstrut & \textbf{Counterfactuals}:\Tstrut \\
    & 1) the carrot was a average. & 1) the baked potatoes was a very sweet and a small thing. \\
    & 2) the only real german food. & 2) great hotel at this. \\
    & 3) the only offer of the mid $\langle unk\rangle$ houses in the last area. & 3) the mashed potato was good. \\
    & 4) the desserts were very bland. & 4) they had the certain menu. \\
    & 5) the hashbrowns were bland. & 5) the good donuts are the food. \\ 
    
    & \textbf{Most frequent words in factuals}: & \textbf{Most frequent words in counterfactuals}:\\
    & within (0.067), num (0.067), days (0.067) & atrocious (0.083), dvd (0.083), used: (0.083)
    \\[0.2em] \hline
    \\[-1em] \multirow{2}{*}[-3em]{\textbf{ABELE}} & 
    \textbf{Saliency}: the desserts \hlRed[0.05]{were} very \hlRed[0.34]{bland}. &
    \textbf{Extrinsic words}: \hlBlue[0.36]{good}\\
    & \textbf{Factuals}: & \textbf{Counterfactuals}: \\
    & 1) the desserts were very bland. & 1) the desserts were very good. \\
    & 2) the desserts were very bland. & 2) the desserts were very good either. \\
    & 3) the desserts were very bland. & 3) the desserts were very good either. \\
    & 4) the desserts were very bland. & 4) the desserts were very good either. \\
    & 5) the desserts were very bland. & 5) the desserts were very good either.
    \\[0.2em] \hline
    \\[-1em] \multirow{2}{*}[-3em]{\textbf{\gbased{}}} & 
    \textbf{Saliency}: the desserts \hlRed[0.10]{were} very \hlRed[0.43]{bland}. &
    \textbf{Extrinsic words}: \hlBlue[0.21]{good} \\
    & \textbf{Factuals}: & \textbf{Counterfactuals}:\\
    & 1) the ingredients were very bland. & 1) the desserts were very good.\\
    & 2) the desserts were very very bland. & 2) the deserts were also very tasty.\\
    & 3) the sausage were very good. & 3) the beers were very good.\\
    & 4) the desserts were very clean. & 4) the desserts were pretty good.\\
    & 5) the sides were very bland. & 5) the rooms were very good.
    \\[0.2em] \hline
    \\[-1em] \multirow{2}{*}[-3em]{\textbf{\pbased{}}} & 
    \textbf{Saliency}: the desserts \hlRed[0.09]{were} very \hlRed[0.33]{bland}. &
    \textbf{Extrinsic words}: \hlBlue[0.43]{great} \hlBlue[0.42]{delicious} \hlBlue[0.38]{excellent} \\
    & \textbf{Factuals}: & \textbf{Counterfactuals}: \\
    & 1) the desserts were very bland. & 1) the desserts were very great . \\
    & 2) desserts were the very bland. & 2) the desserts are very delicious too bland. \\
    & 3) the desserts were bland. & 3) the desserts were very good. \\
    & 4) desserts were very bland & 4) great fresh food, friendly service, bland and were the desserts to go \\
    & 5) were the very bland: desserts & 5) the were very excellent desserts.
    \\[0.2em] \hline
\end{tabular}
\label{tbl:quality}
\end{table*}

The first instance is the review ``\textit{excellent sta for beginning french students.}'' from the Amazon dataset, which BERT correctly classifies as \textit{positive}.
According to the saliency maps, all competitors agree that ``excellent'' is the most influential word.
Similar to the generator-based approaches, \pbased{} extends the insight into the decision-making process by providing the extrinsic word ``\textit{ridiculous}'', illustrating that the classifier learns to leverage sentimental words as evidence for its decisions.
Another extrinsic word ``\textit{but}'', combined with the second counterfactual given by \pbased{}, reveals the model's ability to comprehend contrastive relationships.
The emphasized part in this counterfactual, following the term ``\textit{but}'', outweighs the positive sentiment and dominates the inference.

The second instance, ``\textit{the dessert is very bland.}'', from the Yelp dataset, is assigned a \textit{negative} label by the LSTM model.
Similar to the observation in the first example, XSPELLS delivers limited information with generic words and instances.
LIME identifies the term ``\textit{bland}'' as the most contributing feature, but the notably low importance score poorly matches the model's confidence in its decision.
Meanwhile, the last three approaches concur that the term ``\textit{bland}'' is the primary factor affecting the decision.

Regarding instance-level explanations, the \mbox{(counter-)}factuals listed in the second example underscore the word-level explanations by showing how the presence of ``\textit{bland}'' drives different model decisions.
However, approaches like ABELE and XSPELLS, which rely on na\"{i}ve latent space sampling, produce neighborhoods that lack either diversity or locality, which can be also observed in both examples.
On the other hand, the instances determined by \pbased{}~and XPROAX concentrate more on the target context, effectively complementing the word-level explanations.
The synthetic texts enrich the understanding of the underlying decision-making process when comparing factuals with counterfactuals that share similar components.
With the meaningful texts revealing explanatory information comparable to those from generator-based solutions, both examples imply that iterative probability-based editing offers a proper alternative to a black-box generator.

\subsection{Quantitative evaluation}\label{sec:quantitative}

\begin{table*}[tbp]
\caption{Quantitative evaluation following the three Cs}\label{tbl:effectiveness}
\centering
\fontsize{9.5}{11}\selectfont
\begin{tabular}{|c|l||c|c|c|c|c|} 
\hline
\multirow{2}{*}{Dataset \& Model}
& \multirow{2}{*}{Explainer} 
& \multicolumn{2}{c|}{\textit{Correctness}} & \textit{Completeness} & \textit{Compactness} & \multirow{2}{*}{\makecell[c]{Time cost (s) \\per entry}}\Tstrut \\
\cline{3-6}
& & $R^2$ score & Fidelity & Confidence drop & AOPC & \Tstrut\\
\hline
\multirow{5}{*}{\makecell[c]{Amazon \& BERT}} 
 & LIME & 0.694 ± 0.197 & 0.867 ± 0.123 & 0.346 ± 0.302 & 0.353 ± 0.246 & \textbf{0.160}\Tstrut\\
 & XSPELLS & - & \underline{1.000 ± 0.000} & 0.096 ± 0.225 & 0.082 ± 0.189 & 11.852\\
 & ABELE & 0.869 ± 0.063 & \textbf{0.942 ± 0.030} & 0.358 ± 0.398 & 0.316 ± 0.365 & 364.359\\
 & XPROAX & \textbf{0.878 ± 0.065} & 0.937 ± 0.035 & 0.583 ± 0.369 & \textbf{0.484 ± 0.329} & 8.972 \\
 & \pbased{} & 0.828 ± 0.070 & 0.929 ± 0.026 & \textbf{0.584 ± 0.345} & 0.482 ± 0.311 & 1.019 \\
\hline
\hline
\multirow{5}{*}{\makecell[c]{Amazon \& LSTM}}
 & LIME & 0.880 ± 0.127 & 0.896 ± 0.110 & 0.343 ± 0.293 & 0.331 ± 0.223 & 
\textbf{0.023}\Tstrut\\
 & XSPELLS & - & \underline{1.000 ± 0.000} & 0.067 ± 0.151 & 0.101 ± 0.163 & 10.674 \\
 & ABELE & 0.874 ± 0.059	& 0.936 ± 0.040 & 0.281 ± 0.293 & 0.273 ± 0.249 & 239.322 \\
 & XPROAX & \textbf{0.912 ± 0.053} & 0.938 ± 0.035 & 0.567 ± 0.285 & 0.437 ± 0.234 & 4.582 \\
 & \pbased{} & 0.906 ± 0.045 & \textbf{0.939 ± 0.025} & \textbf{0.582 ± 0.245} & \textbf{0.477 ± 0.220} & 0.804 \\
\hline
\hline
\multirow{5}{*}{\makecell[c]{Yelp \& BERT}}
 & LIME & 0.514 ± 0.280 & 0.882 ± 0.136 & 0.366 ± 0.467 & 0.502 ± 0.421 & \textbf{0.151}\Tstrut\\
 & XSPELLS & - & \underline{1.000 ± 0.000} & 0.231 ± 0.408 & 0.182 ± 0.333 & 11.852\\
 & ABELE & \textbf{0.872 ± 0.056} & 0.950 ± 0.227 & 0.407 ± 0.474 & 0.328 ± 0.409 & 369.802\\
 & XPROAX & 0.858 ± 0.064 & \textbf{0.970 ± 0.024} & \textbf{0.770 ± 0.401} & \textbf{0.629 ± 0.366} & 8.871 \\
 & \pbased{} & 0.777 ± 0.071 & 0.943 ± 0.022 & 0.760 ± 0.405 & 0.609 ± 0.372 & 1.016 \\
\hline
\hline
\multirow{5}{*}{\makecell[c]{Yelp \& LSTM}}
 & LIME & 0.757 ± 0.222 & 0.954 ± 0.101 & 0.506 ± 0.424 & 0.594 ± 0.293 & \textbf{0.023}\Tstrut\\
 & XSPELLS & - & \underline{1.000 ± 0.000} & 0.288 ± 0.397 & 0.225 ± 0.299 & 10.829\\
 & ABELE & \textbf{0.867 ± 0.063} & 0.941 ± 0.039 & 0.448 ± 0.434 & 0.334 ± 0.346 & 233.105\\
 & XPROAX & 0.864 ± 0.071	& \textbf{0.958 ± 0.041} & \textbf{0.827 ± 0.299} & 0.623 ± 0.278 & 4.508\\
 & \pbased{} & 0.833 ± 0.054	& 0.945 ± 0.021 & 0.822 ± 0.296 & \textbf{0.658 ± 0.288} & 0.815\\
\hline
\end{tabular}
\end{table*}

While the qualitative assessments provide valuable insights into the performances of the selected explainers, they are not scalable to conclude the performance of an explainer in general. 
Moreover, qualitative evaluation can be subjective and reliant on the interpretation of results.
For greater objectivity, this section is dedicated to the quantitative evaluation of explanation quality.
Specifically, the evaluation is structured around the three \textit{C}s of explainability~\cite{silva2018towards, nauta2023anecdotal}:
\begin{itemize}
    \item \emph{Correctness}: Explanations should faithfully represent the behaviors of the target model $f(\cdot)$;
    \item \emph{Completeness}: An explanation should encompass all input components that are influential to a model's decision;
    \item \emph{Compactness}: Explanations should be concise, excluding irrelevant features, to present results efficiently.
\end{itemize}

To assess \textit{correctness}, we report the fidelity~\cite{guidotti2018survey} and the $R^2$ score of a surrogate model, both of which indicate the performance of the surrogate in imitating $f(\cdot)$.
An explanation can be faithful only if the surrogate accurately reflects the behaviors of the target model.
\textit{Completeness} is evaluated using explanation-guided manipulation~\cite{nguyen2018comparing}.
Given an explanation, the process masks out the influential words supporting the original decision while inserting the features opposing it.
During the evaluation, influential features are empirically defined with an attribution threshold of $0.1$; features whose absolute attribution exceeds this threshold are involved in the manipulation.
Explanations that guide manipulations inducing larger confidence drops are favored in terms of completeness.
For \textit{compactness}, we measure an explainer's efficiency in presenting results using the AOPC (area over perturbation curve)~\cite{samek2016evaluating}, which is the cumulative sum of confidence drops during manipulation:
$\text{AOPC}_{\boldsymbol{x}} = \frac{1}{l}\sum_{i=1}^l(f(\boldsymbol{x}) - f(\boldsymbol{x}^{(i)}))$,   
where $\boldsymbol{x}^{(i)}$ represents a variant of $\boldsymbol{x}$ after $i$ manipulation steps, and $l$ is the total number of manipulations.
Unlike the evaluation for completeness, manipulations for compactness are applied sequentially following the descending order of relevant features according to their attribution scores.
Compact explanations that prioritize attributions to relevant features will yield high AOPC scores, and vice versa.

The results of the quantitative evaluation are summarized in Table~\ref{tbl:effectiveness}, presenting the average performance on the classifier test set, along with the standard deviations.
XSPELLS, utilizing a decision tree as a more capable surrogate model, outperforms other methods in terms of fidelity.
The different surrogate choice also explains the absence of its $R^2$ score.
However, the non-linear surrogate built in the latent space obstructs the extraction of explanatory information.
As demonstrated by the qualitative examples, estimating feature contributions with word frequencies can overlook the actual relevant features because of the relatively small number of entries on one leaf.
This limitation is reflected by XSPELLS's relatively modest performance in completeness and compactness, aligning with observations from the qualitative evaluation.

Among the remaining explainers that employ a linear regressor as their surrogate, \pbased{}~achieves competitive performance in terms of correctness.
While pushing the figures towards a level comparable to those of the generator-based solutions, \pbased{}~surpasses LIME, which adopts the simple input perturbation method, across various settings.
The high fidelity implies the quality of the constructed neighborhoods, which stick to the underlying data manifold and emphasize only the relevant local decision boundary, thereby simplifying the imitation task for the surrogate.

In addition to correctness, \pbased{}~excels in the evaluations for completeness and compactness.
Without the support of powerful generators, the proposed approach secures the top performance in two of four cases and in general matches the performance of XPROAX, which considerably outperforms the remaining.
These results align with the qualitative assessments, suggesting that \pbased{}~effectively identifies the most contributing features and accurately determines their attribution scores.
Compared to LIME, \pbased{}~benefits from the neighboring extrinsic words during the manipulation test.
These extrinsic words, which oppose the original decision, simplify the reverse of model decisions.
However, compared with XSPELLS and ABELE, which also introduce extrinsic words, it becomes evident that the concrete choice of extrinsic words should be made thoughtfully and informed by insights into the local decision context of the explicand.
Furthermore, while matching the performance of the best performing generator-based solution, \pbased{}~largely reduces time costs as reported in the last column.
The efficiency gain also originates from the substitution of the generative component with editing based on in-text context, highlighting its practical advantages.


\subsection{Stability}\label{sec:stability}

\begin{table}[tbp]
    \caption{Templates for creating similar texts}
    \centering
    \begin{tabular}{cl}
        \hline \\[-1.7mm]
        \multirow{5}{*}{\makecell[c]{\textbf{Templates}}} & 
        1. \textlangle\textit{ADJ}\textrangle~\textlangle\textit{NOUN}\textrangle \\
        \\[-2.3mm]
        & 2. Very \textlangle\textit{ADJ}\textrangle~\textlangle\textit{NOUN}\textrangle \\
        \\[-2.3mm]
        & 3. The \textlangle\textit{NOUN}\textrangle~is \textlangle\textit{ADJ}\textrangle\\
        \\[-2.3mm]
        & 4. A very \textlangle\textit{ADJ}\textrangle~\textlangle\textit{NOUN}\textrangle \\
        \\[-2.3mm]
        & 5. This is a very \textlangle\textit{ADJ}\textrangle~\textlangle\textit{NOUN}\textrangle \\
        \\[-2.3mm]
        \hline
    \end{tabular}
    \label{tbl:templates}
\end{table}

\begin{table}[tbp]
    \caption{Choices of adjective and noun}
    \centering
    \begin{tabular}{cl}
        \hline \\[-1.7mm]
        \makecell[c]{\textbf{Negative}\\\textbf{adjectives}} & \makecell[l]{horrible terrible wrong awful disappointed \\ poor bland worst bad cheap}\\ \\[-1.7mm]
        \hline \\[-1.7mm]
        \makecell[c]{\textbf{Positive}\\\textbf{adjectives}} & \makecell[l]{delicious amazing excellent loved fantastic \\ wonderful perfect fresh great best} \\ \\[-1.7mm]
        \hline \\[-1.7mm]
        \makecell[c]{\textbf{Nouns}} & \makecell[l]{bread soup pizza food meal salad drink \\ dessert fish steak}\\ \\[-1.7mm]
        \hline
    \end{tabular}
    \label{tbl:wordTable}
\end{table}

In addition to effectiveness, \textit{stability} is another desired property of explanation methods, requiring explainers to output similar explanations for similar explicands~\cite{carvalho2019machine}.
Instead of exhaustively searching corpora for similar texts, we evaluate stability with test cases created by text augmentation~\cite{rudinger2018gender}.
Table~\ref{tbl:templates} shows the pre-defined templates for test case creation, which reflect similar contexts commonly found in restaurant reviews.
Substituting the placeholders in the templates, denoted by \textlangle\textit{ADJ}\textrangle~and \textlangle\textit{NOUN}\textrangle, with adjective-noun pairs drawn from Table~\ref{tbl:wordTable} generates one test case containing five similar texts.
A stable explainer should then attribute similar scores to an inserted word across the generated examples in one \textit{test case} due to their contextual similarities.

The stability test adopts the \textit{Yelp LSTM} model as the black box.
The concrete word choices (listed in Table~\ref{tbl:wordTable}) for test case generation are specified for the chosen classifier.
These words are selected according to the confidence scores of the predictions when using them as single-word inputs for the black box.
For each sentiment class, ten adjectives with the highest confidence scores for each class are selected, whereas the chosen nouns possess confidence scores close to 0.5, indicating neutrality for both classes.
Enumerating possible adjective-noun pairs yields a total of 200 test cases.

For each test case, containing five generated texts sharing similar contexts, we measure the consistency of an explainer by computing the deviation in attributions to each inserted word.
A lower deviation across similar contexts signifies greater stability in explanation outcomes.
To further quantify the overall stability of an explainer, we average the attribution deviations for adjectives and nouns respectively over all test cases, yielding the \textit{averaged case deviation} $\bar{\sigma}$.
For example, the averaged case deviation for adjectives is computed by:
$\bar{\sigma}_{\mathrm{adj.}}=\frac{1}{\# \mathrm{Cases}} \sum_{w\in \mathrm{adj.}} \sigma_w$.
Analogously, the averaged case mean $\bar{\mu}$ is computed for each word category to assess whether an explainer uncovers the influence of adjectives and the neutrality of nouns in sentiment analysis.

Table~\ref{tbl:stability} presents the results of the stability evaluation, excluding XSPELLS due to its inconsistent frequency counts for intrinsic features.
The figure in the first column represents the averaged case deviation of model outcomes.
The small deviation indicates trivial prediction variation within each test case, thus strengthening the foundation of the stability evaluation.
The results, particularly the averaged attribution deviations in the fourth and sixth columns, showcase \pbased{}'s~excellence.
It demonstrates low deviations for both word categories, indicating that a word receives consistent attribution scores across similar contexts.
The consistency in explanation results is credited to its controlled generation process driven by probability-based editing.
Moreover, while all competitors agree that the adjectives are more influential than the neutral nouns by showing $\bar{\mu}_{\mathrm{adj.}}\gg\bar{\mu}_{\mathrm{noun}}$, \pbased{}~constantly assigns approximately zero attributions to the nouns, accurately reflecting their non-informative nature in sentiment analysis.
Though ABELE also appears stable in attributions to the adjectives, quantified by $\bar{\sigma}_{\mathrm{adj.}}$, we note that a nonnegligible factor in its low deviation is the considerably smaller scale of attribution scores as shown by $\bar{\mu}_{\mathrm{adj.}}$.

\begin{table}[tbp]
\caption{Stability evaluation reported on Yelp LSTM}
\label{tbl:stability}
\centering
\begin{tabular}{|c|c|cc|cc|} 
\hline
\rule{0pt}{8pt} 
\multirow{2}{*}{\makecell[c]{$\bar{\sigma}_{f(\cdot)}$}} & \multirow{2}{*}{\makecell[c]{Explainer}} & \multicolumn{2}{c|}{Adjective} & \multicolumn{2}{c|}{Noun} \\
\cline{3-6}
& & $\bar{\mu}_{\mathrm{adj.}}$ & $\bar{\sigma}_{\mathrm{adj.}}$ & $\bar{\mu}_{\mathrm{n.}}$ & $\bar{\sigma}_{\mathrm{n.}}$ \rule{0pt}{2.ex}\\
\hline
\multirow{4}{*}{0.005} & LIME & 0.441 & 0.109 & 0.043 & 0.062\Tstrut\\
&ABELE & 0.139 & \textbf{0.062} & 0.064 & 0.054 \Tstrut\\
&XPROAX & 0.441 & 0.148 & 0.035 & 0.055 \Tstrut\\
&XPROB & 0.551 & 0.082 & 0.019 & \textbf{0.019} \Tstrut\\
\hline
\end{tabular}
\end{table}

\subsection{Dependency on external resources}\label{sec:dependency}
The main argument of the paper is that the utilization of opaque generators can induce mistrust in derived explanations.
As the last part of the experiments, this section investigates the dependencies of generator-based approaches on these opaque components.
Due to the space limitation, the dependency analysis is conducted for \gbased{}, the best-performing generator-based explainer according to the quantitative evaluation.
The analysis evaluates the explanation quality of the explainer under various generator configurations.
Specifically, we test the performance of \gbased{}~with two generative models, DAAE and VAE~\cite{bowman2016generating}, and alter the generator capacity by adjusting the bottleneck size of each autoencoder.
To ensure a valid comparison, the network structures of the two generators are kept identical.

Table~\ref{tbl:dependency} details confidence drops and AOPC scores of \gbased{}~when coupled with generators of different configurations on the Amazon LSTM model.
When adopting the DAAE, the explanation quality, in terms of both completeness and compactness, positively correlates to the generator's capacity constrained by the bottleneck size.
In contrast, such a correlation is absent with the VAE.
Despite having an identical network structure to the DAAE and differing only in training objectives, the utilization of the VAE results in a significant falloff in explainer performance regardless of the bottleneck size. 
Table~\ref{tbl:fail_example} showcases the impact of the concrete generator choices on explanation quality through a concrete example.
The attributions derived from the neighborhood constructed via the VAE fail to accurately illustrate the true contributing features, and the instance-level explanation significantly deviates from the target context.
The observed discrepancies suggest that inadequate latent space geometry~\cite{zhao2018adversarially}, lacking the locality-preserving property, is a potential cause of the declined performance.
However, the inherent opacity of generators based on neural networks complicates the prevention and resolution of the issue, underscoring the concerns about adopting opaque components in explanatory frameworks.

\begin{table}[tbp]
\caption{Dependency of \gbased{} on generative models}
\label{tbl:dependency}
\centering
\begin{tabular}{|c|c|cc|} 
\hline
\multirow{2}{*}{\textbf{Generator}} & \multirow{2}{*}{\textbf{Bottleneck}} & \multicolumn{2}{c|}{\textbf{Amazon LSTM}}\Tstrut\\
& & Confidence drop & AOPC\Tstrut\\
\hline
\multirow{4}{*}{DAAE} 
& 32 & 0.464 ± 0.341 & 0.373 ± 0.281\Tstrut\\
& 64 & 0.528 ± 0.302 & 0.411 ± 0.249\\
& 96 & 0.557 ± 0.280 & 0.435 ± 0.236\\
& 128 & \textbf{0.567 ± 0.285} & \textbf{0.437 ± 0.234}\\
\hline \hline
\multirow{4}{*}{VAE} 
& 32 & 0.329 ± 0.364 & 0.289 ± 0.303\Tstrut\\
& 64 & 0.268 ± 0.357 & 0.235 ± 0.302\\
& 96 & 0.296 ± 0.358 & 0.264 ± 0.304\\
& 128 & 0.332 ± 0.352 & 0.299 ± 0.294\\
\hline
\end{tabular}
\end{table}

\begin{table}[tbp]
\centering
\caption{A failed case by \gbased{} with VAE}
\begin{tabular}{p{6cm}p{6cm}} 
    \hline \rowcolor{Gray}
    \multicolumn{2}{m{12.5cm}}{\textbf{Input}: stay the hell away from this book \hfill \textbf{Amazon}, \textbf{LSTM} $b(\cdot)$: {\textcolor{red}{0.86}}\Tstrut} \\[0.2em]
    \hline \\[-0.8em]     
    \multicolumn{2}{m{12.5cm}}{\hlRed[0.27]{stay} the \hlRed[0.10]{hell} \hlRed[0.09]{away} from this book \hfill \textbf{\pbased{} ($|X_p|=5k$)}}\\
    \textbf{Factuals}: & \textbf{Counterfactuals}:\Tstrut\\[-0.1em]
    \makecell[l]{ 
    1) stay the hell away for this book\\
    2) from the hell\\
    3) stay away the book !} &
    \makecell[l]{
    1) hell is a great book\\
    2) this from the book away !\\
    3) stay away from this book}
    \\[1.3em] \hline \\[-0.8em] 
    \multicolumn{2}{m{12.5cm}}{\hlRed[0.32]{stay} the \hlRed[0.17]{hell} \hlRed[0.14]{away} from this book \hfill \textbf{\gbased{}~(DAAE-128)}}\\
    \textbf{Factuals}: & \textbf{Counterfactuals}:\Tstrut\\[-0.1em]
    \makecell[l]{ 
    1) stay the night it with this book!\\
    2) stay the hell away for this book\\
    3) stay the critics away in this book} &
    \makecell[l]{
    1) the far side away from this book\\
    2) enjoy the movie away from this book\\
    3) why the hell do schools this book}
    \\[1.3em] \hline \\[-0.8em] 
    \multicolumn{2}{m{12.5cm}}{stay the hell away from this \hlBlue[0.10]{book} \hfill \textbf{\gbased{}~(VAE-128)}}
    \\
    \textbf{Factuals}: & \textbf{Counterfactuals}:\Tstrut\\[-0.1em]
    \makecell[l]{
    1) where s roy with the rest?\Tstrut\\
    2) where is jamie lee s a?\\
    3) where s to the point?} &
    \makecell[l]{
    1) where s roy with my mind\\
    2) why is so many of these songs?\\
    3) where s the new wave}
    \\[1.3em] \hline 
\end{tabular}
\label{tbl:fail_example}
\end{table}

As a comparison, the dependency of \pbased{}~on its external resource $X_p$ is presented in Table~\ref{tbl:dependency_corpus}.
Its evaluation follows the previous scheme and reports performance under various settings of the external corpus, which serves for probability distribution estimation and prototype selection.
As expected, a larger corpus better represents the underlying probability distribution, thus promoting the quality of text generated by probability-based editing.
This results in a positive correlation between the explanation quality and the size of $X_p$.
However, in contrast to the generator-based solution, \pbased{}~exhibits a direct but less sensitive dependency on the volume of the corpus due to its more controllable text generation process.
While the generator in this experiment requires $200k$ entries for training, the proposed approach still maintains competitive performance with a much smaller corpus of size $5k$.

\begin{table}[tp]
\caption{Dependency of \pbased{} on the prototype corpus}
\label{tbl:dependency_corpus}
\centering
\begin{tabular}{|c|cc|} 
\hline
\multirow{2}{*}{\textbf{Corpus size}} & \multicolumn{2}{c|}{\textbf{Amazon LSTM}}\Tstrut\\
 & Confidence drop & AOPC\Tstrut\\
\hline
2k & 0.478 ± 0.287 & 0.385 ± 0.241\Tstrut\\   
5k & 0.543 ± 0.270 & 0.442 ± 0.235\\
10k & 0.557 ± 0.255 & 0.462 ± 0.227\\
20k & 0.582 ± 0.245 & 0.477 ± 0.220\\
40k & 0.584 ± 0.245 & 0.479 ± 0.217\\
80k & \textbf{0.593 ± 0.239} & \textbf{0.489 ± 0.211}\\
\hline
\end{tabular}
\end{table}


\section{Conclusions}\label{sec:conclusion}
In this work, we propose \pbased{}, a model-agnostic attribution method that improves the neighborhood construction process with probability-based editing.
While following the concept of likelihood maximization for generated texts, the editing approach distinguishes itself from opaque generators with transparent text manipulation on prototypes guided by local \textit{n}-gram contexts.
In addition to the desired transparency of the explanation process, the proposed explainer achieves competitive performance in both qualitative and quantitative evaluations, even compared to the best performing competitor supported by a capable generative autoencoder.

In future work, we plan to refine the recursive editing process of \pbased{}.
Currently, the process greedily selects the optimal operation for each word-prototype pair, without considering the impacts on succeeding manipulations.
The greedy strategy can be unsatisfactory during sequential manipulations.
A potential solution to the problem is a regular restructuring of edited prototypes, thereby promoting the consistency of edits.
Additionally, the setting of the explanation task can be generalized to broaden its applicability.
While this paper focuses on binary classification tasks with short texts for simplicity, the proposed explainer can be conveniently adapted to multi-class scenarios.
This adaptation only involves redefining the counterfactual as an instance with a low prediction confidence score for the class predicted for the explicand.
As for handling longer inputs, probability-based editing can split a long text into shorter segments and manipulate each segment individually, as done for short texts.

\section*{Acknowledgement}
Yi Cai and Gerhard Wunder were supported by the Federal Ministry of Education and Research of Germany (BMBF) in the program of “Souver¨an. Digital. Vernetzt.”, joint project “AIgenCY: Chances and Risks of Generative AI in Cybersecurity”, project identification number 16KIS2013.
Gerhard Wunder was also supported by BMBF joint project “6G-RIC: 6G Research and Innovation Cluster”, project identification number 16KISK020K.

\bibliographystyle{IEEEtran}
\bibliography{ref}


\end{document}